\documentclass{article}
\usepackage{ijcai25}

\usepackage{times}
\usepackage{soul}
\usepackage{url}
\usepackage[hidelinks]{hyperref}
\usepackage[utf8]{inputenc}
\usepackage[small]{caption}
\usepackage{graphicx}
\usepackage{amsmath}
\usepackage{amsthm}
\usepackage{booktabs}
\usepackage{algorithm}
\usepackage{algorithmic}
\usepackage[switch]{lineno}

\usepackage{multirow}
\usepackage{subfig}
\usepackage{amsfonts}
\usepackage{bm}

\title{Improving Prediction Certainty Estimation for Reliable Early Exiting\\ via Null Space Projection}

\author{
Jianing He$^1$\and
Qi Zhang$^1$\and
Duoqian Miao$^{1}$\and
Yi Kun$^2$\and
Shufeng Hao$^3$\and \\
Hongyun Zhang$^{1}$\And
Zhihua Wei$^1$\\
\affiliations
$^1$Tongji University\\
$^2$North China Institute of Computing Technology\\
$^3$Taiyuan University of Technology\\
\emails
\{jnhe, zhangqi\_cs, dqmiao, zhanghongyun, zhihua\_wei\}@tongji.edu.cn,\\
kunyi.cn@gmail.com,
haoshufeng@tyut.edu.cn
}

\begin{document}

\maketitle

\begin{abstract}
   Early exiting has demonstrated great potential in accelerating the inference of pre-trained language models (PLMs) by enabling easy samples to exit at shallow layers, eliminating the need for executing deeper layers. However, existing early exiting methods primarily rely on class-relevant logits to formulate their exiting signals for estimating prediction certainty, neglecting the detrimental influence of class-irrelevant information in the features on prediction certainty. This leads to an overestimation of prediction certainty, causing premature exiting of samples with incorrect early predictions. To remedy this, we define an NSP score to estimate prediction certainty by considering the proportion of class-irrelevant information in the features. On this basis, we propose a novel early exiting method based on the Certainty-Aware Probability (CAP) score, which integrates insights from both logits and the NSP score to enhance prediction certainty estimation, thus enabling more reliable exiting decisions. The experimental results on the GLUE benchmark show that our method can achieve an average speed-up ratio of 2.19$\times$ across all tasks with negligible performance degradation, surpassing the state-of-the-art (SOTA) ConsistentEE by 28$\%$, yielding a better trade-off between task performance and inference efficiency. The code is available at \url{https://github.com/He-Jianing/NSP.git}.

\end{abstract}

\section{Introduction}
Recently, the increasing scale of pre-trained language models (PLMs) has hindered their deployment on resource-constrained devices and latency-sensitive applications due to the high computational costs and long inference time. To address this issue, early exiting~\cite{Deebert,Pabee,globalpast,Berxit,Palbert,ConsistentEE,abs-2412-13236,he2025two}, a kind of adaptive inference strategy, has been proposed to accelerate the inference of PLMs.
Specifically, each intermediate layer of PLMs is paired with an internal classifier to provide an early prediction. Once the certainty level of an early prediction reaches the threshold, the forward inference is terminated, bypassing the computation of the following deeper layers. By conducting a sample-wise inference procedure, PLMs can deal with easy samples with shallow layers and handle hard samples with deep layers. This effectively improves the inference efficiency of PLMs without sacrificing accuracy.

Typically, an early exiting method involves devising an exiting signal to estimate the certainty level of early predictions, determining whether samples exit from early layers.
Researchers have developed quite a few exiting signals by measuring and seeking the characteristics of predictions with high certainty levels, including the entropy~\cite{Deebert,abs-2402-05948}, the softmax score~\cite{Righttool}, the patience~\cite{LECO,BADGE}, the energy score~\cite{ELANG}, and the patience-confidence score~\cite{PCEE}. These methods employ heuristic exiting strategies, enabling the direct computation of exiting signals without involving any learning process and accordingly facilitating easy deployment for inference acceleration. 

Notably, existing methods primarily use logits to devise their exiting signals\footnote{Given that probabilities are calculated from logits, typically through operators like softmax, we collectively label methods based on logits or probabilities as logit-based methods in this paper.}.
Unfortunately, these methods exhibit a high Premature Exiting Rate, which suggests a tendency to overestimate prediction certainty based on logits.
It significantly impairs task performance and hinders the model's acceleration.  
This is attributed to the fact that logits contain only class-relevant information in the features while ignoring the detrimental influence of class-irrelevant information on prediction certainty.
Theoretically, any feature $\bm{x}$ can be orthogonally decomposed into $\bm{x} = \bm{x}^W+\bm{x}^{W^\perp}$ (per Figure~\ref{fig:framework_overview}), where $W$ is the column space of the classifier's weight matrix $\bm{W}$, and $W^{\perp}$ is the null space of $W$. 
We can yield that logits (i.e. $\bm{W}^T\bm{x}$) extract class-relevant information (i.e. the feature similarity to each class) from the component $\bm{x}^W$ but fail to capture class-irrelevant information from $\bm{x}^{W^\perp}$ due to $\bm{W}^{T}\bm{x}^{W^\perp}=\bm{0}$.
Notably, $\bm{x}^{W^\perp}$ is closely related to prediction certainty since a large proportion of redundant (class-irrelevant) information in the feature interferes with the model's classification and reduces prediction certainty.  
Nevertheless, logit-based methods completely disregard the detrimental impact of class-irrelevant information on prediction certainty, inevitably resulting in an overestimation of prediction certainty.
The observations raise an intriguing question: Can we enhance prediction certainty estimation by integrating both class-relevant and class-irrelevant information for more reliable exiting decisions?

In this paper, we propose a novel early exiting method based on the Certainty-Aware Probability (CAP) score, which complements the logit-based methods by considering the proportion of class-irrelevant information in the features. 
To this end, we first define the ratio of $\|\bm{x}^{W^\perp}\|$ and $\|\bm{x}\|$ as the NSP (null space projection) score, which provides an estimation of prediction certainty by leveraging the proportion of class-irrelevant information in the features. A higher NSP score indicates a lower level of prediction certainty. 
Then, we introduce the scaled NSP score as a new logit corresponding to a constructed virtual UNK (unknown) class.
After appending this new logit to the original logits, the CAP score is finally defined as the softmax probability corresponding to the constructed UNK class, which can serve as a proxy for prediction uncertainty. The exiting condition is met once the CAP score falls below the threshold. 
From the formation of CAP, it is evident that our method integrates the class-relevant original logits (indicating the feature similarity to each class) with the class-irrelevant NSP score (indicating the proportion of class-irrelevant information in the features) to enhance prediction certainty estimation, enabling more reliable exiting decisions.

Our contributions are summarized as follows:
\begin{itemize}
    \item We reveal that current logit-based early exiting methods neglect the detrimental influence of class-irrelevant information in the features on prediction certainty, leading to an overestimation of prediction certainty and premature exiting of samples with incorrect early predictions.

    \item We define a class-irrelevant NSP score to estimate prediction certainty by considering the proportion of class-irrelevant information in the features.

    \item We propose a novel early exiting method based on the CAP score, which integrates class-relevant logits and the class-irrelevant NSP score to enhance prediction certainty estimation for more reliable exiting decisions.
\end{itemize}
Extensive experiments on the GLUE benchmark verify that our method outperforms the SOTA ConsistentEE~\cite{ConsistentEE} by $28\%$ in model acceleration, with negligible extra computational or storage overhead.
Further analysis confirms the generality of our method across various backbones and demonstrates its effectiveness in enhancing prediction certainty estimation and delivering reliable exiting decisions.

\section{Related Works}
\paragraph{Heuristic Exiting Strategy.}
DeeBERT~\cite{Deebert}, FastBERT~\cite{Fastbert}, Right-Tool~\cite{Righttool}, and E-LANG~\cite{ELANG} employ entropy, softmax score, and energy score to indicate prediction certainty/uncertainty, respectively. The exiting criterion is met when the certainty level exceeds a threshold. PABEE~\cite{Pabee} relies on cross-layer consistency (i.e. patience) to determine exiting, allowing samples to exit once a sufficient number of consecutive classifiers provide the same answer. F-PABEE~\cite{F-PABEE}, LECO~\cite{LECO}, and BADGE~\cite{BADGE} enhance the flexibility of PABEE by introducing softer cross-layer comparison strategies. PCEE-BERT~\cite{PCEE} proposes a hybrid exiting strategy based on entropy and patience to guarantee reliable and flexible early exiting. 
These logit-based methods neglect the detrimental influence of class-irrelevant information on prediction certainty, compromising the reliability of exiting decisions. To remedy this, we propose to consider the proportion of class-irrelevant information in the features for more reliable exiting decisions. 

\paragraph{Learning-based Exiting Strategy.}
BERxiT~\cite{Berxit}, PALBERT~\cite{Palbert}, and ConsistentEE~\cite{ConsistentEE} train neural networks to generate exiting signals. HASHEE~\cite{hashbased} and BE3R~\cite{BE3R} train networks to route each sample (or token) to an appropriate exiting layer, requiring no layer-by-layer exiting judgments. These methods incorporate the learning process to formulate their exiting strategies, introducing additional parameters and training costs. They also depend on intricate parameter tuning, and carefully designed network architectures or training objectives. In contrast, our method employs a heuristic exiting strategy where exiting signals can be computed directly, facilitating easy deployment.

\paragraph{Architecture/Loss of Multi-Exit Networks.}
CascadeBERT~\cite{cascadebert} performs early exiting in multiple cascaded complete models to provide comprehensive representations for accurate predictions. It also introduces a difficulty-aware objective to calibrate the model's predictions. GPFEE~\cite{globalpast} facilitates early predictions by integrating both past and future states. It incorporates imitation loss to train the model to approximate future states based on all past states. LeeBERT~\cite{zhu2021leebert} and GAML-BERT~\cite{GAML} introduce cross-layer distillation objectives to encourage mutual learning among classifiers. LECO~\cite{LECO} enhances the network architecture based on neural architecture search. BADGE~\cite{BADGE} introduces block-wise bypasses to alleviate optimization conflicts among multiple classifiers. DisentangledEE~\cite{Disentangled} introduces adapters to decouple generic language representations from task-specific representations. It also proposes a non-parametric classifier for improvements. Different from these methods that enhance training objectives or network architectures for early exiting models, our method focuses on improving exiting strategies. Combining our method with these orthogonal works would be an intriguing direction for further research.

\section{Background and Motivation}
\subsection{Problem Definition}
Given a BERT-style PLM with $M$ layers, we denote the hidden states at the $m$-th layer as $h^{(m)}$. To enable early exiting on a classification task with $C$ classes during the inference stage, each intermediate layer is coupled with an internal classifier $f_m$ where $m\in\{1,2,\cdots,M-1\}$ to provide an early prediction $p^{(m)}=f_m(h^{(m)})$, i.e., the probability distribution over the $C$ classes. Classifiers in different layers do not share parameters. The exiting criterion is met when the estimated prediction certainty (or uncertainty) of an internal classifier exceeds (or falls below) the threshold $\tau$.

\subsection{Are Exiting Decisions Reliable?}
We utilize two types of error rates to evaluate the reliability of exiting decisions from different aspects:
\begin{itemize}
\item  \textit{Premature Exiting Rate}, which is the frequency of making "exit" decisions when the internal classifier provides an incorrect early prediction.

\item \textit{Delayed Exiting Rate}, which is the frequency of making "continue" decisions when the internal classifier provides the correct early prediction.
\end{itemize}
The exiting decision-making can be formalized as a binary classification task, where correct early predictions are positive samples and incorrect ones are negative. The Premature Exiting Rate thus corresponds to the false positive rate, indicating the model’s failure to identify negative samples (i.e., incorrect early predictions) due to overestimating prediction certainty. A higher rate suggests that the model is prone to prematurely emit samples with incorrect early predictions, ultimately impairing task performance. 
In contrast, the Delayed Exiting Rate corresponds to the false negative rate, indicating the model’s failure to identify positive samples (i.e., correct early predictions) due to underestimating prediction certainty. As its value escalates, the model tends to delay the exiting of samples with correct early predictions, resulting in prolonged inference time. Therefore, exiting decisions are considered reliable only when both the Premature and Delayed Exiting Rates are sufficiently low.

From Figure~\ref{fig:error_rate}, we observe that the current logit-based early exiting methods exhibit a much higher Premature Exiting Rate compared to the Delayed Exiting Rate. This indicates that these methods primarily suffer from an overestimation of prediction certainty, leading to premature exiting of samples with incorrect early predictions. 
We believe this is because logits represent the feature similarity to each class, namely they are class-relevant. 
However, there is class-irrelevant information related to prediction certainty in the feature space that is not contained in logits. This affects the estimation of prediction certainty, leading to sub-optimal exiting decisions.
These observations trigger our further exploration on the missing information from features to logits.

\subsection{The Missing Information in Logits}
Given a classification task with $C$-classes, each internal classifier consists of a fully connected layer with weight $\bm{W} \in \mathbb{R}^{N \times C}$ and bias $\bm{b} \in \mathbb{R}^C$, which transforms the feature $\bm{x} \in \mathbb{R}^N$ to the logits $\bm{l} \in \mathbb{R}^C$, i.e. $\bm{l} = \bm{W}^T\bm{x}+\bm{b}$, thus providing an early prediction $\bm{p}=\text{softmax} (\bm{l})$. To omit the bias term from the calculation of logits, we offset the original feature space by a vector $\bm{o}=(\bm{W}^T)^+\bm{b}$, where $(\cdot)^+$ denotes the Moore-Penrose inverse of the matrix:
\begin{equation}\label{eq:logits}
\bm{l} = \bm{W}^T\bm{x'}=\bm{W}^T(\bm{x}+\bm{o}).
\end{equation}
For convenience, in the rest of this paper, the feature $\bm{x}$ refers to the offset result. Let $W$ denote the column space of $\bm{W}$ and $W^{\perp}$ denote the null space of $W$. The feature $\bm{x}$ can be orthogonally decomposed into $\bm{x} = \bm{x}^W+\bm{x}^{W^\perp}$, where $\bm{x}^W$ and $\bm{x}^{W^\perp}$ are the projections of $\bm{x}$ to $W$ and $W^{\perp}$, respectively.  Note that $\bm{x}^{W^\perp}$ satisfies $\bm{W}^{T}\bm{x}^{W^\perp}=\bm{0}$, which is class-irrelevant. Therefore, Eq.(\ref{eq:logits}) can be rewritten as:  
\begin{equation}\label{eq:logits_decomposed}
\bm{l} = \bm{W}^T\bm{x}=\bm{W}^T\bm{x}^W,
\end{equation}
where each logit ${l_i}$ is the inner product of $\bm{x}^W$ and the class vector $\bm{w_i}$ (the $i$-th column of $\bm{W}$), indicating the feature similarity to the $i$-th class.
Obviously, logits $\bm{l}$ extracts class-relevant information from the component $\bm{x}^W$ using $\bm{W}$ but neglect the component $\bm{x}^{W^\perp}$ that carries class-irrelevant information. 
Indeed, $\bm{x}^{W^\perp}$ is closely related to prediction certainty.
Precisely, given a feature $\bm{x}$, a larger component $\bm{x}^{W^\perp}$ indicates more redundant (class-irrelevant) information in the feature, which can interfere with the model's classification and lead to reduced prediction certainty. 
Based on the above analysis, the detrimental influence of class-irrelevant information on prediction certainty is completely ignored by logit-based methods, which leads to an overestimation of prediction certainty, causing premature exiting of samples.

To remedy this, we define the ratio of the norms of $\bm{x}^{W^\perp}$ and $\bm{x}$ as the NSP (null space projection) score:
\begin{equation}\label{eq:nsp_score}
\text{NSP}(\bm{x}) =\frac{\|\bm{x}^{W^\perp}\|}{\|\bm{x}\|}.
\end{equation}
The NSP score can provide an estimation of prediction certainty by considering the proportion of class-irrelevant information in the features. Geometrically, it equals to the cosine similarity between the feature and its projection to the null space. The NSP score lies between 0 and 1, and a higher value indicates a lower certainty level.

Logits reflect the feature similarity to each class, while the NSP score indicates the proportion of class-irrelevant information in the features, both of which are closely related to prediction certainty. We hypothesize that combining class-relevant logits and the class-irrelevant NSP score could enhance prediction certainty estimation, thus enabling more reliable exiting decisions.
Such a solution is proposed in Section~\ref{sec:methods} by introducing the scaled NSP score as a new logit.

\begin{figure*}[!t]
\centering
\includegraphics[width=0.95\textwidth]{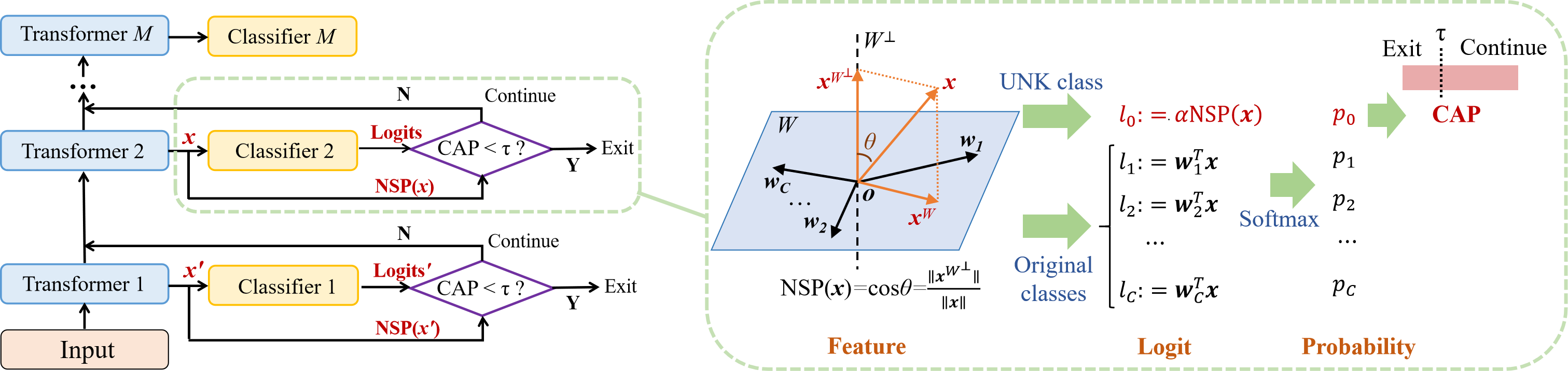}
\caption{Method overview. Our method integrates the class-irrelevant NSP score with class-relevant logits to generate high-quality exiting signals (i.e., the CAP score). The right box details the CAP-based exiting strategy at the second layer. The subspace $W$ is spanned by the class vectors $\bm{w}_1 \sim \bm{w}_C$, i.e., the column vectors of the second-layer classifier’s weight matrix. $W^\perp$ denotes the null space of $W$. Given an offset sample feature $\bm{x}$, $\bm{x}^W$ and $\bm{x}^{W^\perp}$ denote its projections onto $W$ and $W^\perp$, respectively. $\theta$ represents the angle between $\bm{x}$ and $\bm{x}^{W^\perp}$. A scaled NSP score forms a new logit $l_0$ for the virtual UNK class. $\alpha$ aligns the scale of $l_0$ with the original logits $l_1 \sim l_C$ of $C$ original classes. After softmax, the UNK probability $p_0$, i.e., the CAP score, serves as the exiting signal. Exiting occurs if CAP falls below the threshold $\tau$.}
\label{fig:framework_overview}
\end{figure*}

\section{Methods}
\label{sec:methods}
We propose a novel early exiting method based on the Certainty-Aware Probability (CAP) score, which integrates the class-irrelevant NSP score with class-relevant logits for reliable exiting decisions. Figure~\ref{fig:framework_overview} provides an overview of our method. During inference, 
we first compute the NSP score in Eq.(\ref{eq:nsp_score}), which is the ratio of the norms of the feature $\bm{x}$ and its projection $\bm{x}^{W^\perp}$ onto the null space $W^\perp$. Then, we introduce the scaled NSP score as a new logit corresponding to a constructed virtual UNK class. Finally, the exiting signal of our method, i.e. the CAP score, is defined as the softmax probability corresponding to the new logit.

\subsection{Null Space Projection}
This subsection provides the computation of the NSP score. We offset the feature space by $\bm{o}=(\bm{W}^T)^+\bm{b}$ to omit the bias term in the computation of logits, as shown in Eq.(\ref{eq:logits}). Additionally, for computational efficiency, we compute the component $\bm{x}^{W}$, thus obtaining $\|\bm{x}^{W^\perp}\|=\sqrt{\|\bm{x}\|^2-\|\bm{x}^{W}\|^2}$.

As $\bm{x}^{W}$ is the projection of $\bm{x}$ onto the column space of $\bm{W}$, it can be expressed as $\bm{x}^{W}=\bm{W}\bm{\mu}$, where $\bm{\mu} \in \mathbb{R}^C$ denotes the combination coefficients of $\bm{w}_1\sim \bm{w}_C$. Considering that $\bm{x}^{W^\perp}$ satisfies $\bm{W}^{T}\bm{x}^{W^\perp}=\bm{0}$, we have:
\begin{equation}\label{eq:orthogonality}
\bm{W}^{T}(\bm{x}-\bm{x}^{W})=\bm{0}.
\end{equation}
Thus, we construct a system of linear equations regarding the combination coefficients $\bm{\mu}$:
\begin{equation}\label{eq:linear_combination}
\bm{W}^{T}(\bm{x}-\bm{W}\bm{\mu})=\bm{0}.
\end{equation}
Solving Eq.(\ref{eq:linear_combination}), we get $\bm{\mu}=(\bm{W}^T\bm{W})^{-1}\bm{W}^T\bm{x}$, and consequently obtain:
\begin{equation}\label{eq:x_W}
\bm{x}^{W}=\bm{W}(\bm{W}^T\bm{W})^{-1}\bm{W}^T\bm{x}.  
\end{equation}
According to Eq.(\ref{eq:nsp_score}), the NSP score can be computed as:
\begin{equation}\label{eq:nsp_score_xW}
\text{NSP}(\bm{x}) =\frac{\sqrt{\|\bm{x}\|^2-\|\bm{x}^{W}\|^2}}{\|\bm{x}\|}.
\end{equation}

\subsection{The CAP Score}
In this subsection, we introduce the Certainty-Aware Probability (CAP) score, an exiting signal that combines the class-irrelevant NSP score and class-relevant logits. To this end, we introduce the scaled NSP score as a new logit corresponding to a constructed virtual UNK (unknown) class:
\begin{equation}\label{eq:new_logit}
l_0=\alpha\cdot\text{NSP}(\bm{x}),
\end{equation}
where $\alpha$ is the scaling parameter. Note that the NSP score cannot be directly used as a new logit since the subsequent softmax is highly sensitive to the scale of logits.
Hence, we introduce the hyper-parameter $\alpha$ to align the scale of the new logit with each original logit, preventing the final CAP score from being dominated by either side due to scale mismatches.

We append $l_0$ to the original logits $l_1\sim l_C$ and derive an extended probability distribution through softmax:
\begin{equation}\label{eq:prob_distribution}
p_i=\frac{e^{l_i}}{\sum_{i=0}^Ce^{l_i}},
\end{equation}
where $i\in\{0,1,\cdots,C\}$.
Specifically, $l_1\sim l_C$ suggest the feature similarity to each original class, thus the corresponding $p_1\sim p_C$ indicate the probability of the sample belonging to each original class, respectively. In contrast, the new logit $l_0$ reflects the proportion of class-irrelevant information in the feature. 
A larger $l_0$ value indicates a higher proportion of class-irrelevant information in the feature, making it more challenging for the model to identify which class the sample belongs to. Hence, the corresponding $p_0$ denotes the probability of the sample belonging to the constructed UNK class, which indicates the model's inability to provide an exact answer and can serve as a proxy for prediction uncertainty. 
   
Based on the above analysis, the CAP score is defined as:
\begin{equation}\label{eq:cap_score}
\text{CAP}(\bm{x})=p_0=\frac{e^{\alpha\cdot\text{NSP}(\bm{x})}}{\sum_{i=1}^Ce^{l_i}+e^{\alpha\cdot\text{NSP}(\bm{x})}}.
\end{equation}
The CAP score lies in $(0,1)$, which is affected by both class-relevant original logits and the class-irrelevant NSP score. We notice that samples with lower NSP scores (less class-irrelevant information) and larger original logits (higher similarity to each class) typically have lower CAP scores, showing higher prediction certainty. The exiting criterion is met when the CAP score falls below a predefined threshold. Using an information fusion strategy, our method improves prediction certainty estimation to deliver more reliable exiting decisions. The computational overhead from the CAP score is negligible compared to the encoder layer (see Section~\ref{sec:in-depth-analysis}).

\subsection{More Insights into CAP}
\paragraph{Rationale for Logit Concatenation.}
Intrinsically, both the new and original logits capture feature similarity. Specifically, we treat all original classes as known classes, while the constructed virtual UNK class represents all unknown classes. Since the component $\bm{x}^{W^\perp}$ is orthogonal to all known class vectors, it can be naturally interpreted as the class vector for the UNK class. 
Accordingly, the NSP score captures the cosine similarity between the feature and the UNK class vector (per Figure~\ref{fig:framework_overview}), while the original logits measure the feature's similarity to each known class vector.
This semantic alignment justifies concatenating the new and original logits, ensuring that the extended logits and probability distribution carry clear and coherent physical meaning.

\paragraph{Effectiveness of CAP.}
Current logit-based methods only consider the feature's similarity to known class vectors, disregarding the possibility of the sample belonging to an unknown class. This leads to an overly optimistic estimation of prediction certainty.
In contrast, our CAP score considers the feature's similarity to both known and unknown classes, enabling a more objective estimation of prediction certainty.


\section{Experiments}

\subsection{Datasets}
We evaluate our method on six classification tasks of the GLUE benchmark~\cite{glue}, including SST-2, MRPC, QNLI, RTE, QQP, and MNLI. Data statistics are shown in Table~\ref{tab:data_statistics}.

\begin{table}[!t]
\centering
\scalebox{0.75}{
\begin{tabular}{lllll}
\toprule
Dataset & Classes &  $|$Train$|$  & $|$Test$|$ &  Task \\
\midrule
SST-2 & 2 &  67k &  1.8k & Sentiment \\
MRPC & 2 &  3.7k &  1.7k & Paraphrase \\
QQP & 2 &  364k &  391k & Paraphrase \\
MNLI & 3 &  393k &  20k & NLI \\
QNLI & 2 &  105k &  5.4k & QA/NLI \\
RTE & 2 &  2.5k &  3k & NLI \\
\bottomrule
\end{tabular}}
\caption{\label{tab:data_statistics}
    Dataset Statistics. NLI is the Natural Language Inference task, and QA is the Question Answering task.
}
\end{table}

\subsection{Backbones and Baselines}
We choose the widely used BERT-base~\cite{bert} as the backbone model for convincing comparison. We compare our method with two groups of representative and competing baselines. The first group encompasses all existing heuristic exiting strategies, including DeeBERT~\cite{Deebert}, Right-Tool~\cite{Righttool}, PABEE~\cite{Pabee}, PCEE-BERT~\cite{PCEE}, E-LANG~\cite{ELANG}, and F-PABEE~\cite{F-PABEE}. The second group encompasses competitive early exiting methods relevant to our study, including BERxiT~\cite{Berxit}, PALBERT~\cite{Palbert}, DisentangledEE~\cite{Disentangled}, and ConsistentEE~\cite{ConsistentEE}. For fair comparisons, HASHEE~\cite{hashbased} is not included since it employs token-level early exiting, whereas our method focuses on sentence-level early exiting.

For the first group of baselines, DeeBERT, E-LANG, and Right-Tool employ entropy, energy scores, and softmax scores as exiting signals, respectively. 
PABEE uses cross-layer consistency (i.e. the patience score) to determine exiting.  
F-PABEE combines PABEE with softer cross-layer consistency measures to ensure flexible early exiting. 
PCEE-BERT adopts a hybrid exiting strategy based on entropy and patience.  
For the second group of baselines, BERxiT introduces learning-to-exit networks to generate exiting signals that score the certainty level of early predictions. 
PALBERT adopts a deterministic Q-exit criterion that evaluates the cumulative distribution function of the exiting layer’s probability distribution from neural networks for exit decision-making. 
DisentangledEE incorporates adapters to decouple generic language representations from task-specific language representations and further proposes a non-parametric classifier for improvements. It employs the patience-based exiting strategy in PABEE during inference. 
ConsistentEE introduces policy networks to predict the probability distribution of exiting decisions.

\subsection{Experimental Settings}
\paragraph{Speed Measurement.}
Following previous studies~\cite{PCEE,ConsistentEE}, we evaluate the model acceleration with saved layers:
\begin{equation}\label{eq:speedup-ratio}
\text{Speed-up Ratio}=\frac{\sum_{m=1}^M M\times N^m}{\sum_{m=1}^M m\times N^m},
\end{equation}
where $M$ is the total number of layers and $N^m$ is the number of samples exiting from the $m$th layer.
%

\paragraph{Training.}
Our implementation is based on Hugging Face's Transformers~\cite{huggingface}. All internal classifiers are jointly trained with the backbone by minimizing the sum of their cross-entropy losses. Following previous studies~\cite{Pabee,PCEE}, we perform a grid search over learning rates of \{1e-5, 2e-5, 3e-5, 5e-5\}, and batch sizes of $\{16, 32, 128\}$. The maximum sequence length is fixed at $128$. We employ a linear decay learning rate scheduler and the AdamW optimizer~\cite{LoshchilovH19}. We conduct experiments on two RTX4090 GPUs with 24GB.

\paragraph{Inference.}
Following previous studies~\cite{PCEE,F-PABEE}, the batch size for inference is set to $1$, which mimics a common industry scenario where requests from different users arrive sequentially. We select $\alpha$ in Eq.(\ref{eq:new_logit}) from $\{0.01,0.1,1.0,10.0\}$ for each task. 

\begin{table*}
\centering
\scalebox{0.75}{
\begin{tabular}{lccccccc}
\toprule
Method &  RTE  & MRPC  &  QQP  &  SST-2  &  QNLI  &  MNLI & AVG\\
\midrule
BERT-base & 66.4 (1.00$\times$) &  88.9 (1.00$\times$)  & 71.2 (1.00$\times$) &  93.5 (1.00$\times$) & 90.5 (1.00$\times$) & 84.6 (1.00$\times$) & 82.5 (1.00$\times$)\\
\midrule
DeeBERT$^\dag$ & 64.3 (1.95$\times$) &  84.4 (2.07$\times$)  & 70.4 (2.13$\times$) &  90.2 (2.00$\times$) & 85.6 (2.09$\times$) & 74.4 (1.87$\times$) & 78.2 (2.02$\times$) \\
Right-Tool$^\ddag$ & 64.6 (1.92$\times$) &  84.2 (2.04$\times$)  &70.5 (2.04$\times$) &  89.3 (1.92$\times$) & 86.2 (1.96$\times$) & 77.6 (2.04$\times$) & 78.7 (1.99$\times$)\\
PABEE$^\dag$ & 64.0 (1.81$\times$) &  84.4 (2.01$\times$)  & 70.4 (2.09$\times$) &  89.3 (1.95$\times$) & 88.0 (1.87$\times$) & 79.8 (2.07$\times$) & 79.3 (1.97$\times$)  \\
BERxiT & 65.7 (2.17$\times$) &  86.2 (2.27$\times$)  & 70.5 (2.27$\times$) &  91.6 (2.86$\times$) & 89.6 (1.72$\times$) & 82.1 (2.33$\times$) & 81.0 (2.27$\times$)\\    
PCEE-BERT$^\ddag$ & 67.1 (1.89$\times$) &  86.4 (2.13$\times$)  & 70.9 (1.96$\times$) &  92.3 (1.92$\times$) & 88.8 (2.17$\times$) & 82.2 (1.80$\times$) & 81.3 (1.98$\times$)\\
E-LANG$^\ddag$ & 67.2 (1.96$\times$) &  87.0 (1.98$\times$)  &71.0 (1.89$\times$) &  92.2 (2.05$\times$) & 89.6 (1.89$\times$) & 83.0 (1.96$\times$) & 81.7 (1.96$\times$)\\
F-PABEE$^\ddag$ & 67.3 (1.85$\times$) &  87.5 (2.16$\times$)  &70.7 (1.92$\times$) &  92.3 (1.96$\times$) & 89.2 (2.14$\times$) & 82.2 (2.08$\times$) & 81.5 (2.02$\times$)\\
PALBERT$^*$ & 64.3 (1.48$\times$) &  -  &- &  91.8 (1.48$\times$) & 89.1 (1.48$\times$) & 83.0 (1.48$\times$) & -\\
DisentangledEE & 66.8 (1.25$\times$) &  -  &- &  92.9 (1.25$\times$) & 88.5 (1.25$\times$) & 83.0 (1.25$\times$) & -\\
ConsistentEE & \bf{69.0 (1.85$\times$)} &  \bf{89.0 (1.59$\times$)}  & 70.4 (1.82$\times$)$^\ddag$ &  92.9 (1.85$\times$) & 89.9 (1.72$\times$) & 83.4 (1.45$\times$) & 82.4 (1.71$\times$)\\
\bf{Ours}  & 68.6 (1.93$\times$) &  88.0 (2.68$\times$)  & \bf{71.2 (2.07$\times$)} &  \bf{93.0 (2.15$\times$)} & \bf{90.2 (2.40$\times$)} & \bf{83.4 (1.92$\times$)} & \bf{82.4 (2.19$\times$)}\\
\bottomrule
\end{tabular}}
\caption{\label{tab:main_results}
     Performance comparison on the GLUE test set with BERT-base as the backbone. $\dag$ and $*$ denote the results taken from GPFEE \protect\cite{globalpast} and DisentangledEE \protect\cite{Disentangled}, respectively. $\ddag$ denotes the results based on our implementation. Other baseline results are from their original papers. We report F1-score for MRPC and QQP, and accuracy for other tasks. The - denotes unavailable results. Best performance values are marked in bold.}
\end{table*}

\subsection{Overall Performance Comparison}
\label{sec:main_results}
  
In Table~\ref{tab:main_results}, we report the test results of each early exiting method on the GLUE benchmark with BERT-base as the backbone model. The speed-up ratio is approximately $2.00\times$. Overall, our method outperforms all baseline methods in nearly all tasks, which demonstrates the effectiveness of our design. Notably, compared to the backbone, our method achieves an average speed-up ratio of 2.19$\times$ across all tasks with negligible performance degradation, surpassing the SOTA baseline ConsistentEE by 28$\%$. 
  
\begin{figure}[!t]
\centering
\subfloat[SST-2] {\includegraphics[width=0.50\linewidth]{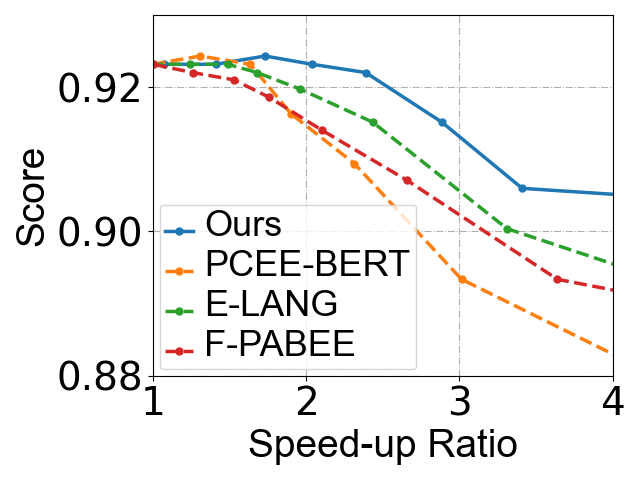}} 
\hfill
\subfloat[QNLI] {\includegraphics[width=0.50\linewidth]{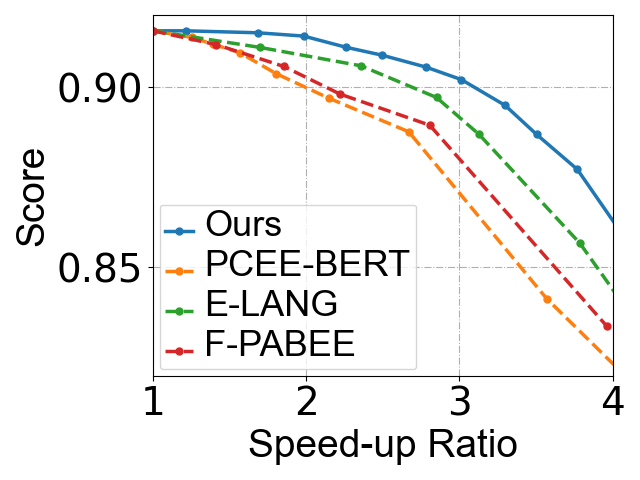}} 
\hfill
\subfloat[MNLI] {\includegraphics[width=0.50\linewidth]{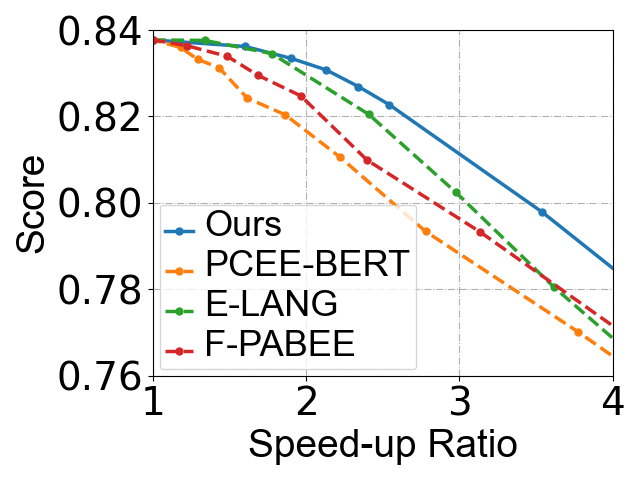}} 
\hfill
\subfloat[QQP] {\includegraphics[width=0.50\linewidth]{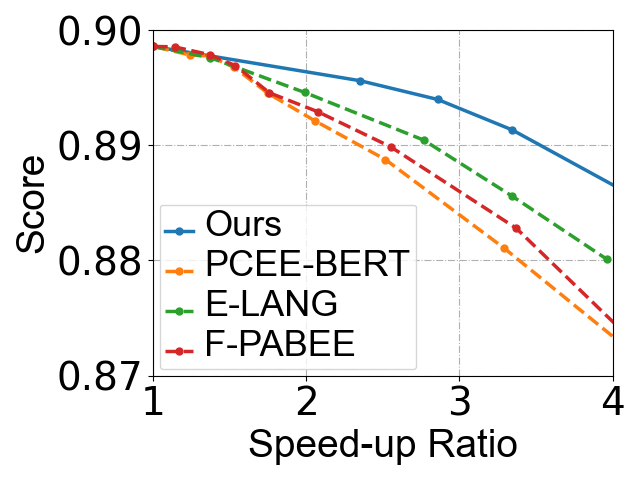}} 
\caption {Performance-efficiency trade-off curves of different early exiting methods on four GLUE development sets.}
\label{fig:trade_off_curve}
\end{figure}

Figure~\ref{fig:trade_off_curve} shows the performance-efficiency curves of our method and three competitive logit-based baselines on a representative subset of GLUE using the same fine-tuned multi-exit network. Our method exhibits a superior trade-off between task performance and inference efficiency across different tasks compared to the baseline methods. 
This demonstrates that our method effectively completes the current logit-based methods by considering the proportion of class-irrelevant information in the features suggested by NSP scores.
Further analysis confirms our method's advantage in estimating prediction certainty and making reliable exiting decisions (see Section~\ref{sec:in-depth-analysis}), offering an intrinsic explanation for its performance gains.

\section{In-depth Analysis}
\label{sec:in-depth-analysis}

\paragraph{DIS Analysis for Exiting Signals.}
Following previous work~\cite{cascadebert}, we use the Difficulty Inversion Score (DIS) to evaluate the quality of prediction certainty estimation.
DIS measures the consistency between the exiting signal and the sample difficulty. A higher value indicates a stronger capability of exiting signals in estimating prediction certainty.
Table~\ref{tab:dis_analysis} shows the DIS of various exiting signals on the SST-2 and QNLI development sets. We observe that by incorporating the class-irrelevant NSP score, the CAP score consistently outperforms the baselines across different layers, demonstrating a more accurate estimation of prediction certainty.
This is crucial for making reliable exiting decisions.

\begin{table}
\centering
\scalebox{0.75}{
\begin{tabular}{l|ccc|ccc}
\toprule
\multirow{2}{1.0cm}{Method}   &  \multicolumn{3}{c}{SST-2} \vline  &  \multicolumn{3}{c}{QNLI}  \\
&  $L=2$ & $L=6$&  $L=10$& $L=2$& $L=6$& $L=10$   \\
\midrule
PCEE-BERT & 66.3 &  73.2  & 75.2 &  54.5 & 65.9 & 70.7 \\
F-PABEE & 71.2 &  78.3  & 81.0 &  55.6 & 68.3 & 71.6 \\
E-LANG & 72.5 &  78.8  & \bf{82.8} &  57.3 & 75.1 & 76.0 \\
\bf{CAP (ours)} & \bf{76.9} &  \bf{83.0}  & 81.1 &  \bf{61.9} & \bf{80.9} & \bf{79.3} \\
\bottomrule
\end{tabular}}
\caption{\label{tab:dis_analysis}
     DIS analysis for each exiting signal at different layers.} 
\end{table}
  
\begin{figure}[!t]
\centering
\subfloat[SST-2] {\includegraphics[width=0.5\linewidth]{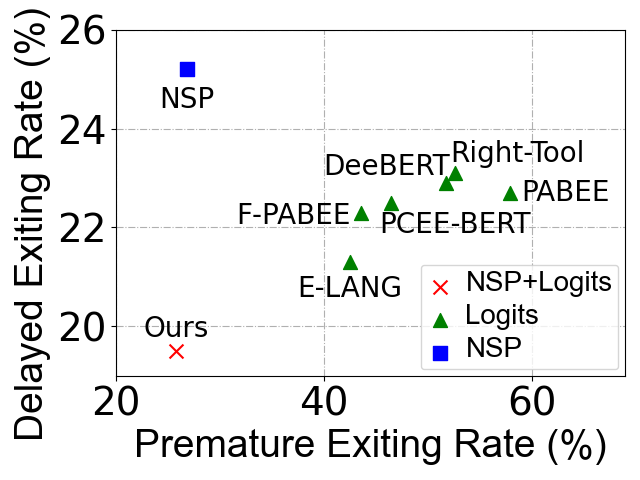}}
\hfill
\subfloat[QNLI] {\includegraphics[width=0.5\linewidth]{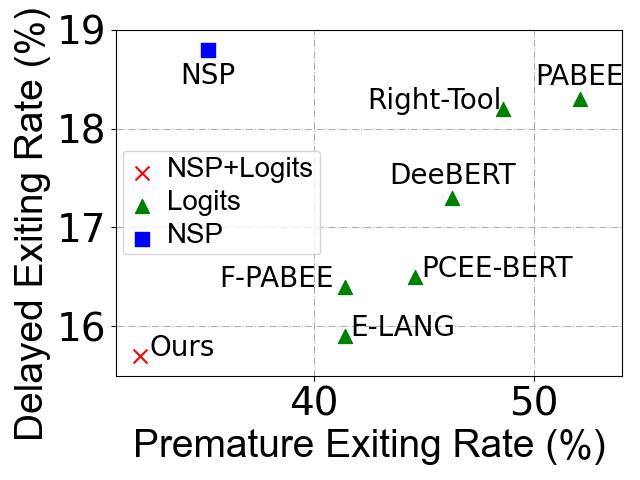}}  
\caption {Two types of error rates for exiting decisions using different early exiting methods under a $4.00\times$ speed-up ratio.}
\label{fig:error_rate}
\end{figure}

\paragraph{Statistics of Exiting Decisions.}
Figure~\ref{fig:error_rate} illustrates the two error rates for exiting decisions using different early exiting methods on the SST-2 and QNLI development sets. 
Compared to the current logit-based methods, utilizing the NSP score as the exiting signal can effectively reduce the Premature Exiting Rate while exhibiting a relatively high Delayed Exiting Rate. We attribute this to the NSP score’s neglect of feature similarity to each class, resulting in an underestimation of prediction certainty and delayed exiting for samples with correct early predictions.
In contrast, by fusing information from both class-relevant logits and the class-irrelevant NSP score, our method significantly reduces both types of error rates, demonstrating more reliable exiting decisions.
This explains the superiority of our method in model acceleration.

\begin{figure}[!t]
\centering
\subfloat[SST-2] {\includegraphics[width=0.50\linewidth]{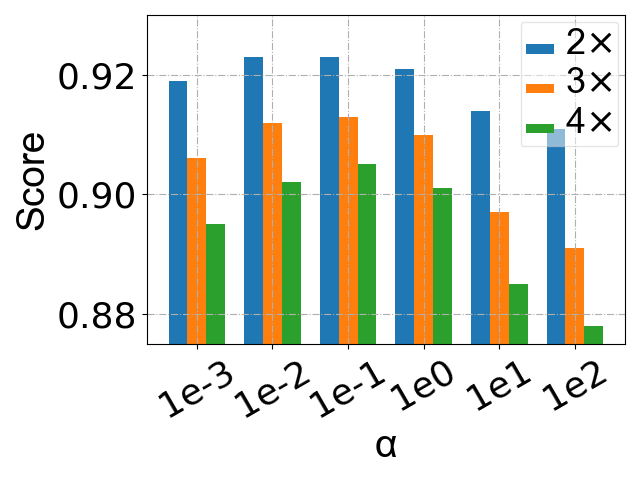}} 
\hfill
\subfloat[QNLI] {\includegraphics[width=0.50\linewidth]{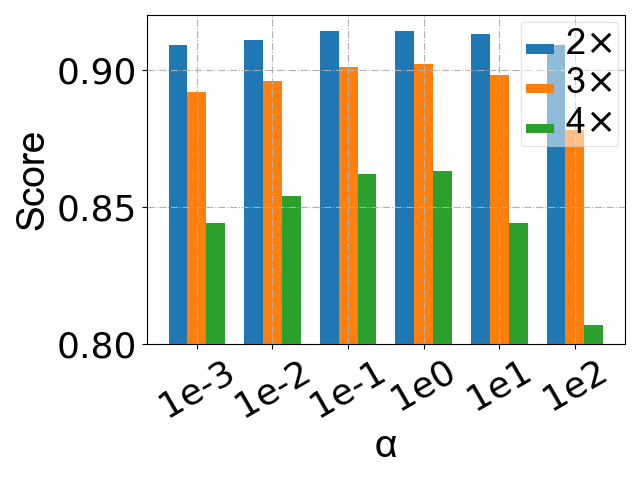}} 
\caption {Impact of $\alpha$ on the task performance under different speed-up ratios for SST-2 and QNLI tasks.}
\label{fig:pa_alpha}
\end{figure}

\paragraph{Impact of $\alpha$.}
Figure~\ref{fig:pa_alpha} shows the impact of $\alpha$ in Eq.(\ref{eq:new_logit}) on model acceleration. $\alpha$ is used to align the scale of the new and original logits, and a higher value indicates a stronger emphasis on the class-irrelevant NSP score compared to the class-relevant original logits. 
Overall, we observe that both excessively large and small values of $\alpha$ can impair the model acceleration under various speed-up ratios. This suggests an optimal trade-off between class-irrelevant and class-relevant information, which enables reliable exiting decisions. Additionally, the incorporation of the NSP score yields more significant performance improvements under high acceleration scenarios. 
We attribute this to the limited classification capacity of shallow classifiers. This aggravates the overestimation of prediction certainty when relying solely on the class-relevant original logits, exacerbating premature exiting of samples with erroneous early predictions.
Hence, incorporating the class-irrelevant NSP score becomes increasingly crucial for reliable exiting decisions. Finally, for parameter selection, values of $\alpha$ between $0.1$ and $1.0$ typically deliver satisfactory results across different tasks and speed-up ratios.

\paragraph{Computational and Storage Costs.} 
Table~\ref{tab:computational_costs} presents the computational complexity for each module in our model. It is noteworthy that, at each layer, the offset vector $\bm{o}$ in Eq.(\ref{eq:logits}) and the projection matrix in Eq.(\ref{eq:x_W}) are shared across samples during inference, requiring no repetitive calculations. Consequently, for each sample, the computational overhead introduced by making exiting decisions primarily arises from incorporating internal classifiers and calculating CAP, totaling less than 1.21M FLOPs per layer. This overhead is negligible compared to the 1813.5M of an encoder block, leading to a minimal impact on inference time. The inference time analysis presented in Table~\ref{tab:inference_time} further reinforces this conclusion.
Additionally, the results in Table \ref{tab:storage_costs} indicate that our model only requires less than $0.03\%$ additional parameters compared to the backbone due to incorporating internal classifiers.
The analysis above confirms our method’s efficiency regarding computation and storage, suggesting strong scalability to larger datasets and backbones.
Furthermore, as the model's computational complexity is primarily dominated by encoder blocks, its inference costs are approximately proportional to the number of executed layers, validating the rationale for the speed measurement in Eq.(\ref{eq:speedup-ratio}).

\begin{table}[!t]
\centering
\scalebox{0.75}{
\begin{tabular}{lcc}
\toprule
\multirow{2}{3.0cm}{Module} & \multicolumn{2}{c}{FLOPs} \\
 & $C=2$ & $C=3$ \\
\midrule
Embedding & 786.4K & 786.4K\\
Encoder & 1813.5M & 1813.5M\\
Pooler & 1.2M & 1.2M\\
Classifier & 3.1K & 4.6K\\
Offset Vector$^*$ & 4.4K & 8.4K\\
Projection Matrix$^*$ & 1.8M & 3.0M\\
CAP Calculation$^*$ & 1.2M & 1.2M\\
\bottomrule
\end{tabular}}
\caption{\label{tab:computational_costs} Analysis of computational complexity. $C$ denotes the number of classes. $*$ signifies the modules introduced by our method.}
\end{table}

\begin{table}[!t]
\centering
\scalebox{0.75}{
\begin{tabular}{ccc}
\toprule
Model & Total Inference Time  &  Overhead vs. BERT-base\\
\midrule
BERT-base & 8.36s &  +0.0\%\\
DeeBERT & 8.73s &  +4.4\%\\
Right-Tool & 8.60s &  +2.9\%\\
PABEE & 8.68s &  +3.8\%\\
E-LANG & 8.81s &  +5.4\%\\
Ours & 8.90s &  +6.4\%\\
\bottomrule
\end{tabular}}
\caption{\label{tab:inference_time}
Comparison of total inference time in the SST-2 development set. Each model disables early exiting by setting an unreachable threshold to guarantee full-layer execution. Our method incurs acceptable overhead compared to standard early exiting baselines.}
\end{table}

\begin{table}[!t]
\centering
\scalebox{0.75}{
\begin{tabular}{lcc}
\toprule
\multirow{2}{3.0cm}{Model} & \multicolumn{2}{c}{\#Params} \\
 & $C=2$ & $C=3$ \\
\midrule
BERT-base & 109.48M & 109.48M\\
Ours & +16.92K & +25.38K\\
\bottomrule
\end{tabular}}
\caption{\label{tab:storage_costs}
    Comparison of parameter volumes. 
}
\end{table}

\paragraph{Generality on Different PLMs.} 
To verify the generality of our information fusion strategy, we apply our method to ALBERT~\cite{albert}, which is an optimized version of BERT with reduced parameters and improved efficiency. The results at around $2.00\times$ speed-up ratio are listed in Table~\ref{tab:generality}. Our method still outperforms all competitive baseline methods in general, validating its generality on various PLMs.

\begin{table}[!t]
\centering
\scalebox{0.75}{
\begin{tabular}{lllllll}
\toprule
Method &  Speed-up &  QQP  &  SST-2  &  QNLI  &  MNLI & AVG\\
\midrule
ALBERT-base$^\dag$ & 1.00$\times$ & 79.6  &  93.3  & 92.0  & 85.2  & 87.5 \\
\midrule
PABEE$^\dag$ & 1.95$\times$ &  \bf{79.8}  &  92.4  & 90.9  & 84.2  & 86.8 \\
PALBERT & 1.21$\times$ &  79.1  &  91.4  & 90.9  & 83.2  & 86.2 \\
DisentangledEE & 1.26$\times$ &  79.3  &  92.2  & 91.0  & 83.5  & 86.5 \\
\bf{Ours}  & 2.11$\times$ &  79.4  &  \bf{92.8}  & \bf{91.5}  & \bf{84.8}  & \bf{87.1} \\
\bottomrule
\end{tabular}}
\caption{\label{tab:generality}
     Test results on a representative subset of GLUE with ALBERT-base as the backbone. The speed-up ratio is averaged across 4 tasks. We report the mean of accuracy and F1-score for QQP, and accuracy for other tasks.
     $\dag$ denotes results taken from GPFEE \protect\cite{globalpast}. Other baseline results are taken from DisentangledEE \protect\cite{Disentangled}. } 
\end{table}

\section{Conclusion}
In this paper, we propose a novel early exiting method based on the CAP score, which integrates insights from both class-relevant logits and the class-irrelevant NSP score to address the overestimation of prediction certainty in current logit-based early exiting methods, enabling more reliable exiting decisions.
Our method is simple yet effective. Extensive experiments on the GLUE benchmark validate its superiority across different backbones. Further analysis confirms the interpretability of our method and its efficiency in terms of storage and computation.

\clearpage

\appendix
\section*{Acknowledgments}
This work was supported by the National Natural Science Foundation of China (No.~62376198, No.~62406225), the National Key Research and Development Program of China (No.~2022YFB3104700), and the Shanghai Baiyulan Pujiang Project (No.~08002360429).

\section*{Contribution Statement}
Duoqian Miao serves as the corresponding author and is responsible for all communications related to this manuscript.

\bibliographystyle{named}
\bibliography{ijcai25}

\end{document}